\documentclass[11pt]{article}

\usepackage[preprint]{acl}

\providecommand{\iflatexml}{\iffalse}
\iflatexml
  \makeatletter
  \AtBeginDocument{\let\@maketitle\@empty}
  \makeatother
\fi

\usepackage{times}
\usepackage{latexsym}

\usepackage[T1]{fontenc}

\usepackage[utf8]{inputenc}

\usepackage{microtype}
\usepackage{inconsolata}

\usepackage{graphicx}
\usepackage{booktabs, makecell,xspace,subcaption}
\usepackage{amsmath,amssymb}
\usepackage{array}
\usepackage[dvipsnames]{xcolor}

\usepackage[most]{tcolorbox}
\tcbuselibrary{listingsutf8}
\usepackage{listings}

\newcommand{\UpArrow}[1][ForestGreen]{\textcolor{#1}{\((\uparrow)\)}}
\newcommand{\DownArrow}[1][BrickRed]{\textcolor{#1}{\((\downarrow)\)}}

\newcommand{\dtitle}{\textsc{MIMIC-Ext-SynNotes}\xspace}

\definecolor{darkgray}{rgb}{0.225, 0.225, 0.225}
\definecolor{varcolor}{rgb}{0.4, 1.0, 1.0}
\definecolor{bracecolor}{rgb}{0.6, 0.8, 1.0}

\lstdefinestyle{mypython}{
    backgroundcolor=\color{white},
    basicstyle=\ttfamily\footnotesize\color{black},
    keepspaces=true,
    breaklines=true,
    breakatwhitespace=false,
    breakindent=0pt,
    breakautoindent=false,
    prebreak=\mbox{},
    postbreak=\mbox{},
    columns=fullflexible,
    showstringspaces=false,
    frame=none,
    language={},
    keywordstyle=,
    stringstyle=,
    commentstyle=,
    moredelim=**[is][\color{varcolor}]{@@}{@@},
    moredelim=[s][\color{bracecolor}]{\{\{}{\}\}},
    literate={{@}{}{1}},
}

\newtcblisting{promptbox}[1][Custom Title]{%
    title=\texttt{#1},
    listing only,
    listing options={style=mypython},
    colback=white,
    colframe=gray!60,
    colbacktitle=gray!30,
    coltitle=black,
    fonttitle=\bfseries,
    enhanced,
    frame hidden=false,
    boxrule=0.5pt,
    arc=1mm,
    left=2mm,
    right=2mm,
    top=1mm,
    bottom=1mm
}

\title{Systematic Evaluation of the Quality of Synthetic Clinical Notes \\ Rephrased by LLMs at Million-Note Scale}

\author{
 \textbf{Jinghui Liu\textsuperscript{1}}\quad\quad
 \textbf{Sarvesh Soni\textsuperscript{2}}\quad\quad
 \textbf{Anthony Nguyen\textsuperscript{1}}
 \\
 \textsuperscript{1}Australian e-Health Research Centre, CSIRO, Australia\\
 \textsuperscript{2}National Library of Medicine, National Institutes of Health, USA
 \\
 \nolinkurl{jinghui.liu@csiro.au}
}

\begin{document}
\maketitle

\begin{abstract}

Large language models (LLMs) can generate or synthesize clinical text for a wide range of applications, from improving clinical documentation to augmenting clinical text analytics. Yet evaluations typically focus on a narrow aspect -- such as similarity or utility comparisons -- even though these aspects are complementary and best viewed in parallel. In this study, we aim to conduct a systematic evaluation of LLM-generated clinical text, which includes intrinsic, extrinsic, and factuality evaluations of synthetic clinical notes rephrased from MIMIC databases at million-note scale. Our analysis demonstrates that synthetic notes preserve core clinical information and predictive utility for coarse-grained tasks despite substantial linguistic changes, but lose fine-grained details for task like ICD coding. We show this loss of detail can be substantially mitigated by rephrasing notes by chunks rather than by the whole note, but at the cost of reduced factual precision under incomplete context. Through fact-checking and error analysis, we further find that synthesis errors are dominated by misinterpretation of clinical context, alongside temporal confusion, measurement errors, and fabricated claims. Finally, we show that the synthetic notes -- despite their task-agnostic nature -- can effectively augment task-specific training for rare ICD codes. 

\end{abstract}

\section{Introduction}

Large language models (LLMs) have demonstrated impressive capabilities in generating human-like text, including text written by domain experts~\cite{Shao2024-ly}. In healthcare, studies have shown that LLMs can generate clinical notes similar to those drafted by clinicians~\cite{Van_Veen2024-ko,Small2025-ma,Kahl2026-xb}, driving interest in LLM-assisted clinical note-taking to reduce the significant burden of documentation~\cite{Roberts2024-ou,Colicchio2019-fs}. 
The flexible mechanism of generative LLMs also naturally lends them to synthetic data generation, where LLMs can be prompted to produce task-specific, synthetic clinical texts to help provide larger training samples for improving model performance and fairness in healthcare~\cite{Juwara2024-ij,Ktena2024-zw}, especially since real-world, private patient records are often difficult to access.

Given the growing adoption of LLMs in clinical text generation, it is essential to measure the quality of the generated text. Studies on LLM-generated clinical documentation typically compare the generated notes with human-written reference notes via an array of textual similarity metrics~\cite{Croxford2025-tw,Carandang2025-cg}, which measure the intrinsic quality of the LLM generation. For studies using LLMs for data synthesis, the evaluation is often performed based on a specific downstream task~\cite{Loni2025-pq,Alshaikhdeeb2025-ss}, which focuses on the extrinsic utility of the generated content. While exceptions exist~\cite{Wang2024-iv}, these different evaluation types are rarely presented jointly on a consistent set of synthetic clinical notes.  Since existing metrics typically capture only a limited aspect of textual quality, a systematic evaluation combining multiple strategies, spanning multiple metrics and downstream tasks, is vital to understand how LLMs generate clinical text.

In this study, we aim to address this gap through a multi-angle evaluation of LLM-generated clinical notes, including 1) \textbf{similarity-based intrinsic evaluation}, 2) \textbf{utility-based extrinsic evaluation}, 3) \textbf{fairness evaluation in downstream applications}, 4) \textbf{fact-checking with LLM-as-judge}, and 5) \textbf{human analysis of LLM errors}. We apply these setups to clinical notes rephrased by LLMs, which we treat as synthetic notes. 

Rephrasing can be viewed as the most basic form of text generation given that LLMs are provided with complete input contexts, and serves as a lean test case of whether LLMs can interpret clinical information in its original form. Furthermore, LLMs are considered particularly proficient in rephrasing noisy input into high-quality text, a technique now used widely in curating large-scale web corpora~\cite{Maini2025-os,Li2024-ew}. 
Intuitively, LLMs should be able to rephrase clinical notes with ease, but our findings reveal that significant information loss can occur, especially when rephrasing whole notes compared to smaller chunks --- a finding demonstrated and cross-checked across our evaluation setups.

Additionally, we examine generation scenarios with varied prompts and repeated generation, and discuss their implications.
Although the rephrased synthetic notes are task-agnostic, we show they can be valuable resources for data augmentation in a task-specific setting --- demonstrated through ICD prediction~\cite{Ji2024-ca,Gan2025-ge}. The benefits are particularly substantial for modeling rare ICD codes. 
These experiments shed light on the strengths and limitations of LLM-generated clinical notes.

\section{Related Work}

Evaluation is key to driving the development of models in NLP. In the clinical domain, proper evaluation is critical to ensure an NLP system is fit-for-purpose and can be applied in healthcare, where both accuracy and safety are critical. For LLM-generated text, existing works usually evaluate it by comparing to the human-written counterparts. For example, studies on generating clinical documentation using LLMs~\cite{Chung2025-xp,Soni2024-ex} --- such as drafting a discharge summary based on patient records~\cite{Schwieger2024-tb} or writing a report from a medical image~\cite{Yu2023-ac,Nicolson2024-rr} --- typically compare the generated notes with ground-truth notes written by human experts. This often involves multiple textual similarity metrics, such as ROUGE~\cite{Lin2004-ip} and BERTScore~\cite{Zhang2020-ao}. Crafting pairwise similarity metrics is a research field in itself~\cite{Croxford2025-tw}. 

Another application of LLM-based text generation is clinical text synthesis~\cite{Mitra2025-jw,Liu2024-oz,Li2023-pt}, which provides data augmentation for a given task. These studies typically focus on particular note types or labeled datasets, and perform evaluation based on the downstream task of interest~\cite{Tang2023-hb,Soni2020-ld}, with task accuracy serving as the extrinsic measure. Unlike the aforementioned works, leveraging downstream tasks sidesteps the need for expert-written reference notes that are costly to produce at scale. 

It is known that many existing metrics for text evaluation are imperfect~\cite{Sai2022-wg} and only capture a specific aspect of textual features. Therefore, it is common to aggregate various metrics when comparing and ranking the performance of different generation models, and such aggregation tends to correlate better with human judgment~\cite{Xu2024-gd,Van_Veen2024-ko}. Furthermore, combining and analyzing different evaluation approaches can better illuminate generation quality by allowing results to be compared and cross-checked.
In this study, we conduct such a systematic, multi-angle evaluation on a consistent set of synthetic notes generated at large scale --- a combination that, to our knowledge, is still lacking in prior work.

Clinical text is noisy and varies in quality, a characteristic that resembles web data. Curating web data is an important research direction given its key role in training language models~\cite{Li2024-ew}, and this process can benefit from LLM-based synthesis. LLMs can be instructed to generate content with high ``educational value''~\cite{Penedo2024-rz,Gunasekar2023-vr} or to produce improved versions of existing text snippets~\cite{Maini2024-zd,Kang2025-vv}. The latter is also known as \textit{rephrasing}. It has been found that even small-sized LLMs ($<$8B) are capable of producing high-quality rephrased corpora for improved and more efficient pretraining, and that larger LLMs do not necessarily improve downstream pretraining quality~\cite{Kang2025-vv,Maini2024-zd}. This method has also been shown effective in the medical domain~\cite{Liu2024-iq} for domain-specific pretraining compared to text synthesis from patient profiles alone~\cite{Kweon2024-xk}. Given its effectiveness, general applicability, and natural pairing with human-written notes for intrinsic evaluation, we adopt this method to generate synthetic clinical notes for our multi-angle evaluation.

\section{Synthetic Clinical Notes by Rephrasing the MIMIC Databases}
\label{sec:method}

We generate synthetic clinical notes using LLMs by grounding on human-written notes, which are sourced from the MIMIC databases. We describe the synthesis process in the following.

\subsection{Data Preparation}
The MIMIC databases provide de-identified, real-world clinical data across various modalities. We considered all clinical notes from MIMIC-III (v1.4)~\cite{Johnson2016-kd} and MIMIC-IV (v2.2)~\cite{Johnson2023-qk}. 
MIMIC-IV includes two note types (i.e., discharge summary and radiology report) while MIMIC-III provides a broader range of documentation charted during hospital encounters (e.g., physician and nursing notes). We align the format of the de-identification placeholders to underscores (``$\_\_\_$''), consistent with MIMIC-IV. After deduplicating notes with the exact same content, the final pool of MIMIC notes comprises \textit{4.6M clinical notes} with \textit{1.4B words}. 

\subsection{Rephrasing Setup}

Two approaches were considered in rephrasing clinical notes. The first is to rephrase \textit{by the entire note}, preserving the original context. 
The second splits long notes into chunks and rephrases them \textit{by chunk}. This is motivated by findings that LLMs may struggle to process information as the input context grows longer~\cite{Liu2024-pa,Hsieh2024-nw}, including during rephrasing~\cite{Maini2024-zd}. Each chunk contains approximately 150 words with sentence boundaries maintained, resulting in an average of $2.3$ chunks per MIMIC note. 

Small-scale generative LLMs are capable of effective rephrasing while maintaining efficiency~\cite{Kang2025-vv,Maini2024-zd}. Following this approach, we use three \textasciitilde8B models to rephrase MIMIC notes into ``high-quality'' text: Llama-3.1 (8B)~\cite{Dubey2024-xd}, Mistral-v0.3 (7B)~\cite{Jiang2023-zm}, and Qwen-2.5 (7B)~\cite{Qwen2024-bw}. These efficient open-source LLMs also represent a realistic deployment choice for privacy- and resource-constrained clinical environments. Prompt templates, decoding parameters, and other inference details are provided in Appendix~\ref{sec:generation}. The complete process required about 1.8K GPU hours on NVIDIA H100 with vLLM~\cite{Kwon2023-cu}, producing a \textbf{9.1B-word} corpus, which we denote as \dtitle (details in Appendix Table~\ref{tab:synthetic}).

\section{Intrinsic Evaluation and Comparison with Human-written Notes}
\label{sec:intrinsic}

\begin{figure}[h!]
  \centering
  \includegraphics[width=.9\linewidth]{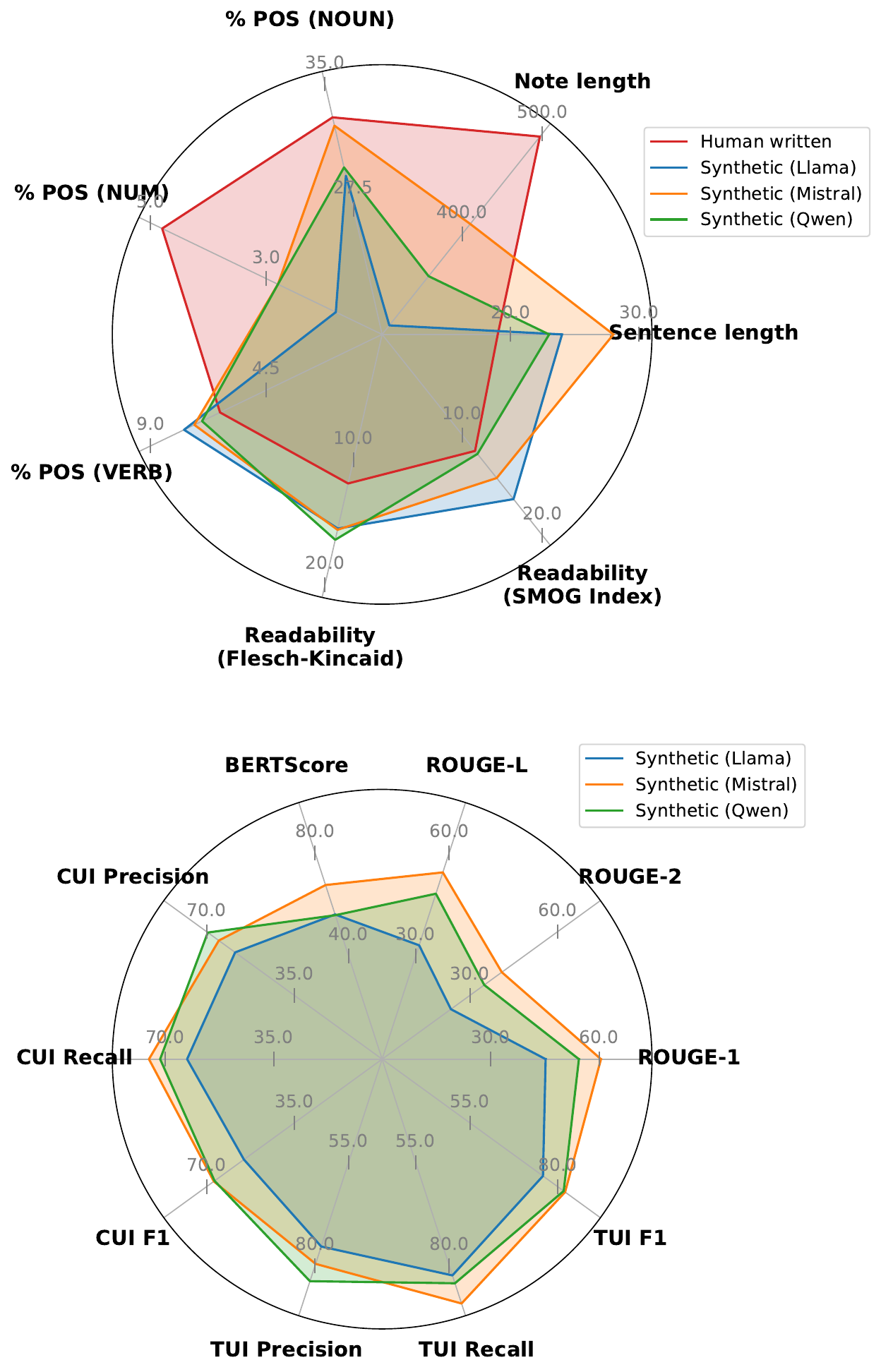}
  \caption{Linguistic features of synthetic notes and their textual similarities to human-written notes.}
  \label{fig:intrinsic}
\end{figure}

\begin{figure*}[t!]
  \centering
  \includegraphics[width=1.0\linewidth]{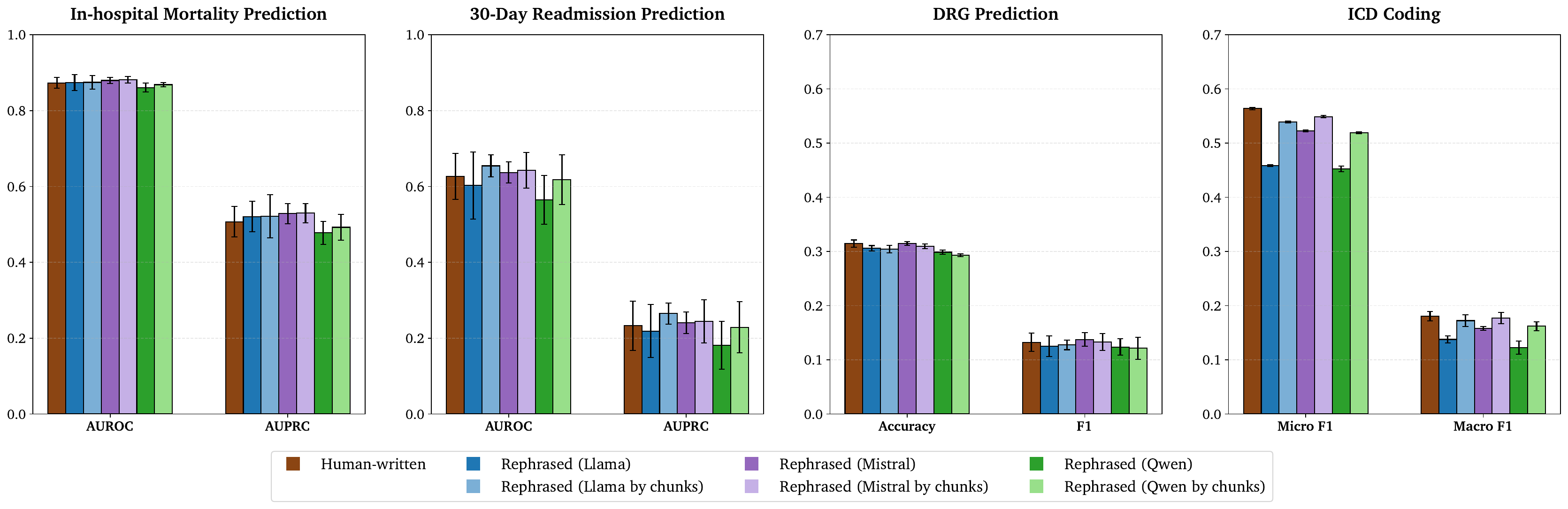}
  \caption{Comparing synthetic notes with human-written notes in downstream modeling. 
  Each metric is calculated as the mean of five runs, and the error bars represent 95\% confidence intervals.}
  \label{fig:extrinsic}
\end{figure*}

\subsection{Experimental Setup}
\label{sec:eval-intrinsic}

For intrinsic evaluation, we analyzed and compared surface-level and linguistic features between synthetic and human-written notes. 
We computed sentence and note lengths, part-of-speech (POS) distributions (for nouns, verbs, and numerals), and two standard readability scores: Flesch-Kincaid Grade Level~\cite{Solnyshkina2017-kp} and SMOG Index~\cite{Harry1969-ur}. 

For pairwise metrics to capture the textual similarity between human-written and synthetic notes, we computed ROUGE~\cite{Lin2004-ip} and BERTScore~\cite{Zhang2020-ao}. Due to token length limitations in many pre-trained BERTScore base models (typically capped at 512 tokens), we opted for a DeBERTa-based model~\cite{He2023-wj} with flexible input length and adaptation to natural language inference, which offers enhanced capability for semantic similarity comparison.

To quantify the semantic fidelity of LLM-generated notes in terms of medical concepts, we followed previous works~\cite{Gao2022-dv,Xu2024-gd} to extract UMLS Concept Unique Identifiers (CUIs) using QuickUMLS~\cite{Soldaini2016-xj}, and computed precision, recall, and F1 scores using the original notes as the reference, as proposed in MedCON~\cite{Yim2023-jd}. CUIs capture unique medical concepts from UMLS as a bridge for shared medical semantics. Following MedCON, we only considered clinically relevant semantic groups, such as anatomy, chemicals \& drugs, and disorders. For a high-level comparison, we also mapped CUIs to their medical semantic types (Type Unique Identifiers, or TUIs) and computed the corresponding scores.

To facilitate efficient evaluation, we sampled 1K discharge summaries and 5K other note types from MIMIC-IV and MIMIC-III, respectively, as we observed that using more notes did not yield significantly different results.

\subsection{Results}

Figure~\ref{fig:intrinsic} shows the comparison between human-written and LLM-generated notes. The synthetic notes are noticeably shorter in length but contain longer sentences with higher readability scores. They also show \textbf{a significant shift in lexical and semantic distributions}, as reflected by textual similarity scores with different degrees of variation across LLMs. 

Analysis of TUI distributions (Table~\ref{tab:top5_Added_Omitted}) reveals that the synthetic notes tend to omit concepts related to \textit{``Body Space or Junction''} and \textit{``Medical Device''} while introducing more from \textit{``Natural Phenomenon or Process''} and \textit{``Clinical Attribute''}, suggesting that synthetic text is \textbf{more descriptive but less dense with anatomical and specialist terminology}. 
This shows that LLMs may have a tendency to introduce more general terms and phrases in the generated text. 
The lists of frequent errors in semantic types made by each LLM are presented in Appendix Table~\ref{tab:semtype}.

\begin{table}[h]
\centering
\resizebox{\linewidth}{!}{
\begin{tabular}{llr}
\hline
& \textbf{Semantic Type} & \textbf{\%} \\
\hline
 & Natural Phenomenon or Process & 16.5\% \\
  & Clinical Attribute & 13.2\% \\
 Added & Phenomenon or Process & 9.8\% \\
  & Pharmacologic Substance & 9.5\% \\
  & Organ or Tissue Function & 8.2\% \\
\hline
 & Body Space or Junction & 12.2\% \\
  & Mental Process & 11.8\% \\
 Omitted & Medical Device & 11.6\% \\
  & Gene or Genome & 11.2\% \\
  & Cell Component & 8.8\% \\
\hline
\end{tabular}
}
\caption{Top 5 semantic types added (false positives) or omitted (false negatives) in synthetic notes, aggregated across three LLMs.}
\label{tab:top5_Added_Omitted}
\end{table}

\section{Extrinsic Evaluation with Healthcare Applications}
\label{sec:extrinsic}

\subsection{Experimental Setup}

We compared the utility of synthetic notes with human-written notes by training downstream clinical NLP models as a form of extrinsic evaluation. 
For a patient in a given task cohort, the input notes for the downstream model are either synthetic or human-written for both training and testing. In other words, we compared \textit{``train synthetic, test synthetic'' (TSTS)} and \textit{``train real, test real'' (TRTR)} by only varying the source of notes. 
The downstream performance is considered as a proxy to measure whether synthetic notes provide similar predictive utility to human-written notes. For this purpose, we did not consider \textit{TSTR} (train synthetic, test real) or \textit{TRTS} (train real, test synthetic), which can be confounded by factors such as writing styles and other features in the textual distribution.\footnote{Section~\ref{sec:aug} presents results for data augmentation, which aligns with \textit{TSTR}.}

We considered four clinical NLP tasks that represent key healthcare applications: risk stratification (in-hospital mortality and 30-day hospital readmission prediction)~\cite{Jiang2023-zv,Liang2025-lm} and medical code assignment (DRG and ICD-10 prediction)~\cite{Wang2024-sn,Douglas2025-jg}. On the input side, they involve multiple note types authored by different caregivers at various stages of the care episode: in-hospital mortality~\cite{Jiang2023-zv,Si2019-gi} and DRG prediction~\cite{Wang2024-sn} tasks occur during the hospital episode and thus utilize all available clinical notes charted by the time of prediction (one day after ICU admission), such as nursing notes and radiology reports. The 30-day hospital readmission prediction~\cite{Boag2021-in,Huang2019-cb} and ICD coding~\cite{Ji2024-ca,Liu2022-qi} tasks occur after the end of hospital stay, and their input is based on the discharge summary alone.

On the output side, these tasks also involve varying complexities and requirements for clinical details: the first three tasks require the knowledge of overall patient status for a single-label classification, whereas the ICD coding task requires a higher level of attention to clinical details for the challenging multi-label classification with 8K ICD-10 codes~\cite{Edin2023-zj}. 
The four task cohorts are constructed from MIMIC with sample sizes ranging from 18K to 122K patients, and require processing of 546K notes. The statistics of the datasets are presented in Appendix Table~\ref{tab:cohorts}. 

For a given downstream task and a source of clinical notes, a classification model was trained, evaluated on the validation set for early stopping, and then tested on the held-out set to report final results. The only variation for a given task was the source of clinical notes, which are either human-written or synthetic, i.e., rephrased by one of the LLMs.  
Appendix~\ref{sec:eval-extrinsic} presents the details of the task cohorts and the training of downstream models.

\subsection{Results} 

Figure~\ref{fig:extrinsic} shows that synthetic notes exhibit comparable downstream performance to human-written notes on single-label tasks (i.e., all tasks except ICD coding), with results overlapping the 95\% confidence intervals of using original notes. This suggests that 
\textbf{key clinical information is preserved in synthetic notes}, enabling accurate modeling and prediction for three prediction tasks.
On the other hand, the ICD coding task (8K classes; multi-label) that requires fine-grained details exhibits much lower performance (with a drop of 4--10\% in micro F1 score), suggesting \textbf{loss of certain clinical details} during synthesis that are essential for coding.
Crucially, rephrasing \textit{by chunks} instead of \textit{by notes} substantially closes this performance gap, indicating \textbf{LLMs' difficulty in retaining and integrating information from long contexts} as a primary driver for detail loss.

\subsection{Fairness Evaluation on Risk Stratification}
\label{sec:fairness}

\begin{figure*}[h!]
  \centering
  \includegraphics[width=.75\textwidth]{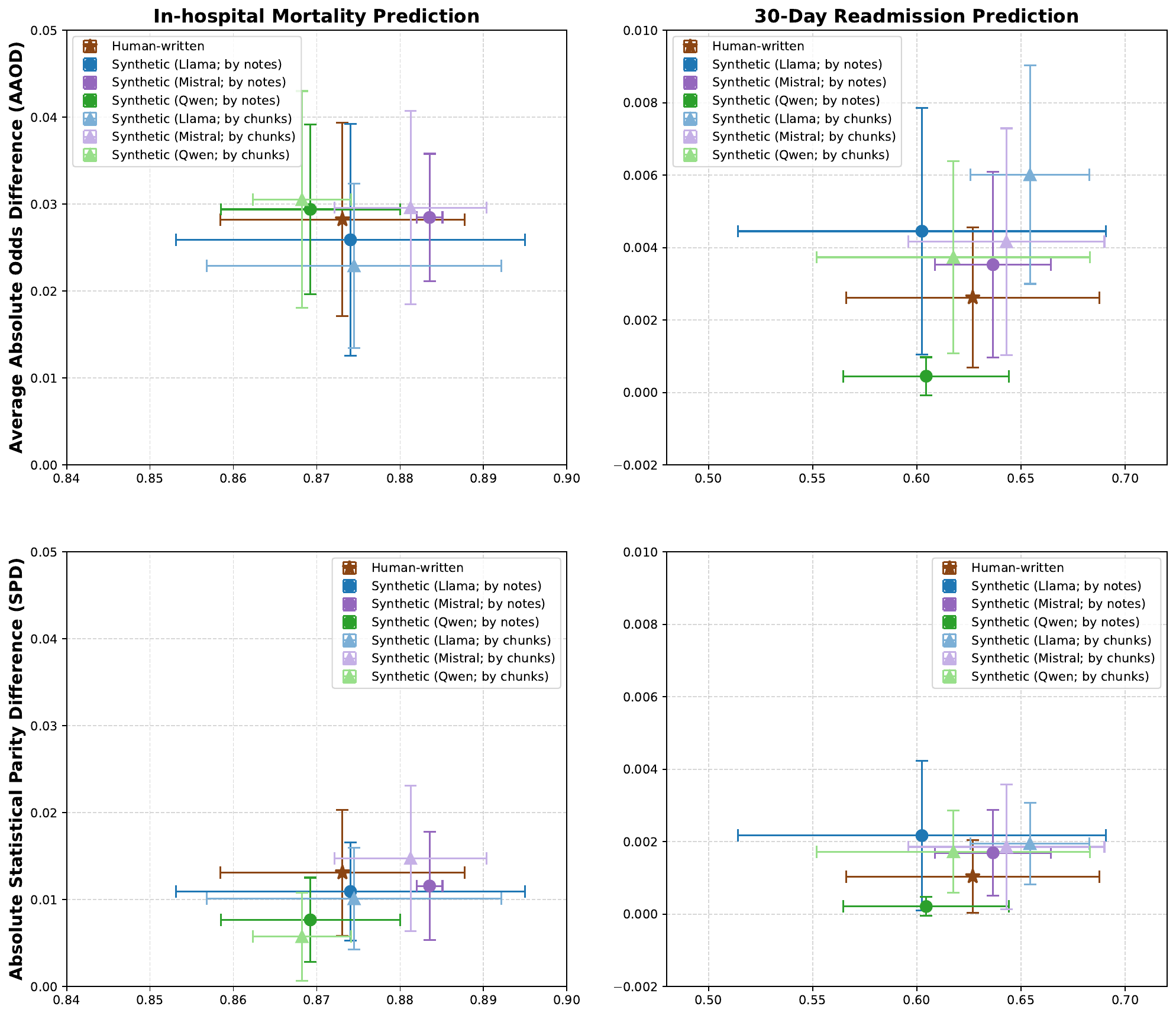}
  \caption{Fairness evaluation of the synthetic notes in downstream modeling in comparison with human-written notes. The x-axis presents model performance (AUROC score) and the y-axis presents model fairness (Average Absolute Odds Difference (AAOD) on the top panel and the absolute Statistical Parity Difference (SPD) on the bottom panel). For fairness metrics, the lower the better. The value of 0 for AAOD and SPD means no model discrimination between the examined subgroups. The results were averaged across age, gender, and ethnicity.}
  \label{fig:fairness}
\end{figure*}

Beyond predictive accuracy, we also assess fairness for two risk stratification tasks (in-hospital mortality and 30-day hospital readmission predictions). 
Following \citet{Brown2025-yj}, we split the patients into subgroups based on age (over and under 50), gender (female and male), and ethnicity (non-white and white), and computed the average absolute odds difference (AAOD)~\cite{Hardt2016-tg} and statistical parity difference (SPD)~\cite{Besse2021-bg} across these subgroups.

Figure~\ref{fig:fairness} presents the results of the fairness evaluation, where we observe no significant changes in the fairness of the models trained on synthetic notes compared to those trained on human-written notes. This indicates that the LLMs did not introduce significant biases in their rephrasing that would affect downstream modeling of risk stratification.

\section{Fact-checking with Error Analysis}
\label{sec:fact}

We measure the truthfulness of the synthetic notes by fact-checking their claims against those made in the original notes. We then analyze the common error patterns exhibited by LLMs.

\subsection{Experimental Setup}

We follow \citet{Munnangi2025-nc} to decompose synthetic notes into atomic facts~\cite{Min2023-qi} and conduct fact-checking via entailment~\cite{Wang2024-hs}. Following~\citet{Xie2024-uc}, we calculate \textit{precision} (the proportion of synthetic factual claims entailed by the human-written note) and \textit{recall} (the proportion of reference factual claims entailed by those in the synthetic note). 

For a pair of synthetic note $N_{S}$ and human-written note $N_{H}$, they are decomposed into atomic factual claims $C_{S}$ and $C_{H}$. Precision is defined as the proportion of facts in $C_{S}$ that can be entailed by $N_{H}$: 

\begin{equation}
    Precision=\frac{1}{|C_{S}|}\sum_{c\in{C_{S}}}[{N_{H}} \vDash {c}]
\end{equation}

where ${N_{H}}\vDash{c}$ means note $N_{H}$ entails fact $c$.
Recall is defined as the proportion of facts in $C_{H}$ that can be entailed by those from the synthetic notes $C_{S}$: 

\begin{equation}
    Recall = \frac{1}{|C_{H}|}\sum_{c\in{C_{H}}}[{C_{S}} \vDash {c}]
\end{equation}

Note that this is different from \citet{Munnangi2025-nc}, which used sentences from the reference note instead of decomposed facts; we found the latter to be more interpretable. The final precision and recall are macro averages over the sampled datasets, and F1 is the harmonic mean of the final precision and recall scores.  
We use Gemini-2.5 to decompose facts and assess entailment due to its strong reasoning capabilities.
Due to the cost of querying the closed-source model to provide robust assessment, we conducted this part of the experiments on a small set of sampled notes as exploratory analysis, which covered more than 10K facts. More details are described in Appendix~\ref{sec:method-fact}.

\subsection{Results} 

Table~\ref{tab:fact} shows that synthetic notes generally exhibit \textbf{high precision in their factual claims, but considerably lower recall}, indicating missing details.
This becomes more pronounced when rephrasing \textit{long notes} (>2K words), with recall as low as 0.3. Meanwhile, rephrasing \textit{by chunk} produces a substantially larger number of facts, improving recall and suggesting a preservation of clinical details, consistent with findings in Section~\ref{sec:extrinsic}. 

Beyond aggregated metrics, one author with a background in medical informatics analyzed sample outputs and identified a taxonomy of common error types (see Appendix Table~\ref{tab:errors}): \textbf{misinterpretation of clinical context is the dominant failure mode}, followed by measurement errors, temporal/recency confusion, and fabrication of claims when the documentation is subtle. 
LLMs sometimes make up names and dates even though these are removed from real notes, and they also struggle with shorthand notation, e.g., confusing blood pressure values for respiration rates.

Finally, the results highlight a trade-off inherent to chunk-level rephrasing: while it maintains more claims and improves recall, it also lowers precision and is thus accompanied by higher hallucination rates.
In many of these cases, LLMs see incomplete context and may generate plausible but unsupported content whose reference context appears in another chunk.

\begin{table}[]
    \centering
    \resizebox{\linewidth}{!}{
    \begin{tabular}{ll ll r}
    \toprule
        & Fact Prec & Fact Rec  & Fact F1 & \# Avg. Facts \\ 
    \midrule
Llama & 0.903 & 0.587 & 0.712 & 68 \\
\cmidrule{2-5}
\ \ \textit{long notes}  & 0.904  & 0.309 & 0.460 & 101 \\
\ \   \textit{by chunk} & 0.868 \DownArrow & 0.794 \UpArrow & 0.830 \UpArrow & 297 \UpArrow \\
\midrule
Mistral & 0.972 & 0.812 & 0.885 & 105 \\
\cmidrule{2-5}
\ \ \textit{long notes} & 0.959 & 0.719 & 0.822 & 186 \\
\ \ \textit{by chunk} & 0.938 \DownArrow & 0.779 \UpArrow & 0.851 \UpArrow & 251 \UpArrow \\
\bottomrule
    \end{tabular}
    }
    \caption{Fact-checking results based on decomposed atomic facts from synthetic notes.}
    \label{tab:fact}
\end{table}

\begin{figure}[h!]
  \centering
    \includegraphics[width=\linewidth]{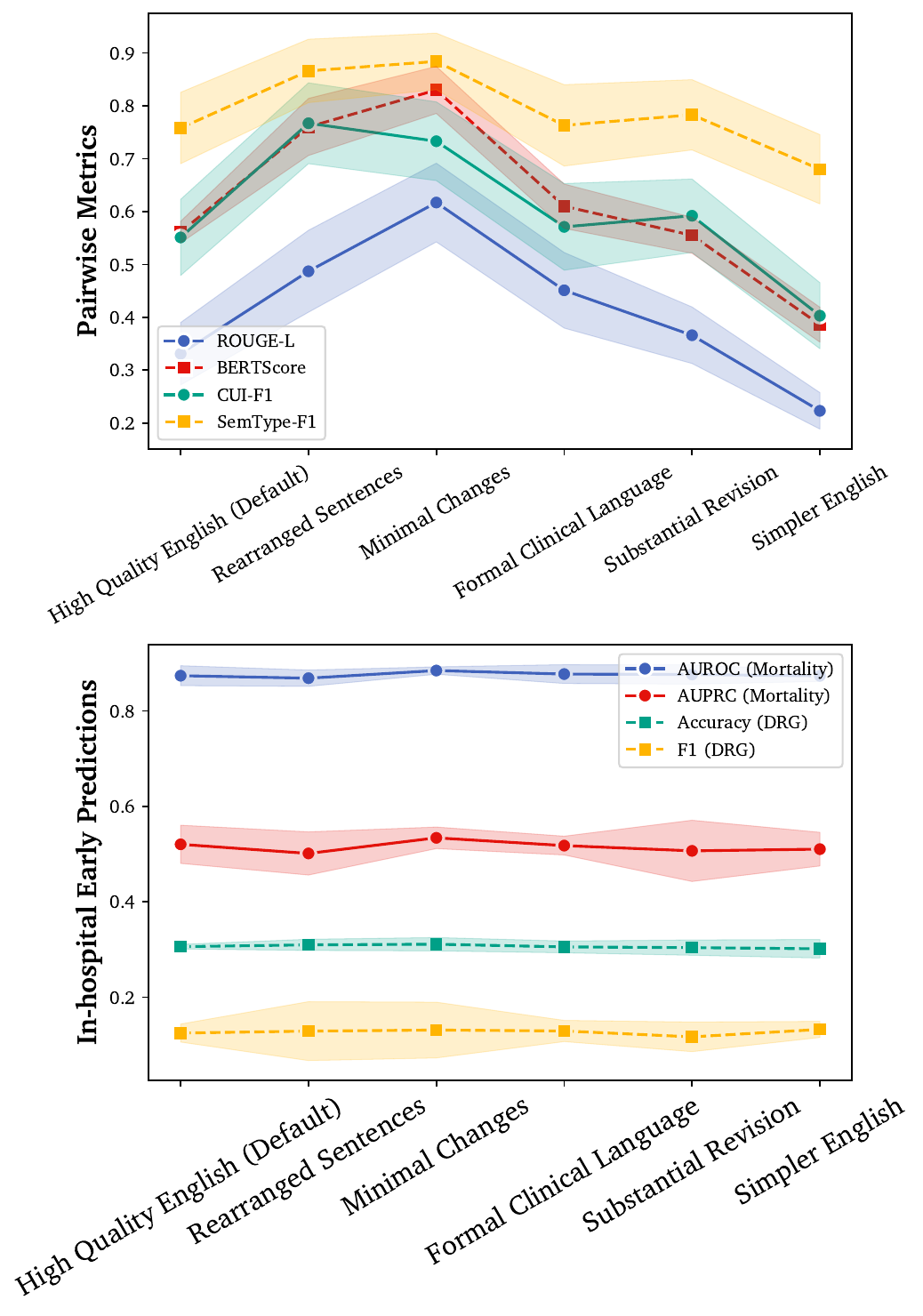}
  \caption{Impact of varied prompt instructions on intrinsic and extrinsic evaluations. Results are averaged over five random seeds.}
  \label{fig:prompt}
\end{figure}

\begin{figure}[h!]
  \centering
    \includegraphics[width=\linewidth]{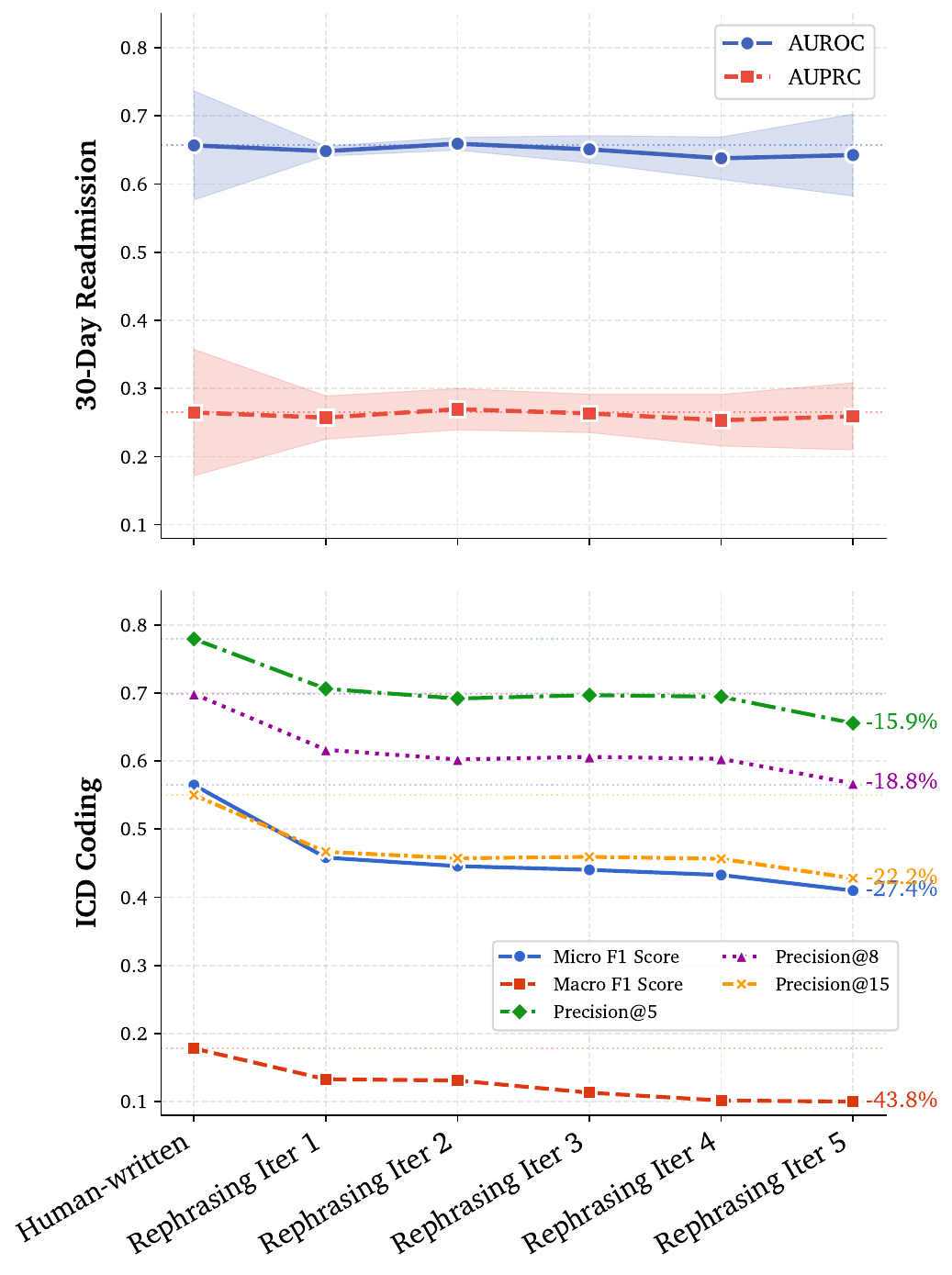}
  \caption{Impact of repeated synthesis on 30-day hospital readmission prediction across 5 iterations. Results are averaged over five random seeds.}
    \label{fig:iteration}
\end{figure}

\section{Additional Analyses}

We conduct additional analyses to examine the impact of prompting setups on the generated notes, including different instructions and iterative prompting. We also perform a data augmentation experiment on the ICD prediction task.

\subsection{Impact of Varied Prompts}
We examine alternative prompts (Appendix~\ref{sec:alter_prompt}) that vary in structure (e.g., \textit{``rearranging sentences''}) and style (e.g., \textit{``simplified English''}), and observe their impact on synthetic notes, reflected by both intrinsic and extrinsic measurements.
As expected, prompt variation significantly affects textual similarity: both syntactic and semantic contents change dramatically from instruction to instruction (Figure~\ref{fig:prompt}; top). Interestingly, the default instruction to produce \textit{``high-quality''} language has a similar effect to \textit{``substantial revision''}, showing the complexity of handling clinical documentation from the LLM's perspective. 

On the other hand, the change of prompts has minimal impact on downstream modeling (Figure~\ref{fig:prompt}; bottom). This indicates that the synthetic notes preserve key clinical information related to patient severity despite differences in linguistic style, consistent with the findings in Section~\ref{sec:extrinsic}.

\begin{table*}[tbh!]
    \centering
    \resizebox{0.8\textwidth}{!}{
\begin{tabular}{l|cc|cc}
\hline
 & \multicolumn{2}{c|}{All codes} & \multicolumn{2}{c}{Rare codes} \\
 & micro F1 & macro F1 & micro F1 & macro F1 \\
\hline
Human-written notes
 & \makecell{0.5644 \\ \small(0.5622--0.5667)}
 & \makecell{0.1738 \\ \small(0.1703--0.1772)}
 & \makecell{0.0359 \\ \small(0.0173--0.0596)}
 & \makecell{0.0134 \\ \small(0.0070--0.0199)} \\[6pt]
\textit{+synthetic notes}
 & \makecell{\textbf{0.5703} \\ \small(0.5671--0.5736)}
 & \makecell{\textbf{0.2141} \\ \small(0.2007--0.2278)}
 & \makecell{\textbf{0.1035} \\ \small(0.0595--0.1485)}
 & \makecell{\textbf{0.0448} \\ \small(0.0258--0.0656)} \\
\hline
\end{tabular}
    }
    \caption{Augmentation results for ICD coding. Models are evaluated on the human-written notes in the test set by bootstrapping 1000 samples from multiple seeds. The rare codes are 391 ICD codes that appeared only 10 times in the cohort (the benchmark in \citet{Edin2023-zj} excluded codes that appeared fewer than 10 times). The reported scores are mean values and 95\% confidence intervals. All augmented results show significant improvements with p-value $<$0.01 with two-sided t-test.}
    \label{tab:aug}
\end{table*}

\subsection{Impact of Repeated Synthesis}
We explore repeated synthesis, where the LLM iteratively rephrases its own outputs to examine semantic drift. \citet{Shumailov2024-hp} demonstrates cases where recursive generations can lead to model collapse when training happens in a loop. This raises critical concerns over the quality of data for training models over time, as generated content gradually becomes an integral part of the data pool, especially after deploying LLMs in practice.

Figure~\ref{fig:iteration} shows the results on two downstream tasks over five iterations of repeated synthesis, each based on the output from the previous round. The performance on the single-label readmission prediction task remains stable.
For ICD coding, although the performance decreases over time and sometimes by a large margin, the decline is progressive rather than sudden, and we have yet to observe a ``collapse'' in downstream modeling, i.e., scoring zero accuracy. 
Note that in this setting, the ``generator'' (i.e., the LLM) was kept unchanged, and only the downstream models (i.e., task-specific ``predictor'') were trained at each iteration instead of the ``generator''~\cite{Shumailov2024-hp}.
Moreover, the provision of a reference serves as a constraint to keep LLMs from generating completely irrelevant content. 
Nevertheless, the issue of LLMs missing fine-grained clinical details persists, as demonstrated by the declining ICD performance over repeated rephrasing.

\subsection{Rephrased Notes for Augmentation}
\label{sec:aug}

We examine the value of the synthetic notes in augmenting the training set for downstream modeling. Using LLMs to generate synthetic training data by following annotated examples has been studied for various clinical NLP tasks~\cite{Tang2023-hb,Suvalov2025-ub}. Unlike these studies, which craft application-specific prompts with demonstrations (i.e., few-shot prompting), our synthesis does not consider any task-related instructions or labels in the prompt. 
In other words, we apply the \textbf{task-agnostic synthetic notes for task-specific data augmentation}.

We focus on the challenging ICD-10 coding task and augment the original training set with synthetic notes generated by Mistral (by chunks), and evaluate on the human-written notes.
Note that this \textit{``test real''} setting is different from Section~\ref{sec:extrinsic}, where synthetic notes were examined in a \textit{TSTS} setting.

The results are presented in Table~\ref{tab:aug}, where we see that augmentation with synthetic notes increases all F1 scores (significant improvements with p-value $<$0.01), especially for rare codes (micro F1: $0.03\rightarrow0.10$). This shows the additional potential of LLM-synthesized clinical text for data augmentation despite being task-agnostic.

\section{Discussion \& Conclusion}

Existing studies on synthetic clinical text typically evaluate the generation from one angle and lack direct comparisons between these evaluation strategies on a consistent set of clinical notes.
We consider these different angles complementary in essence and useful for providing a unified view of the synthetic text quality; thus, we included textual similarity comparison, downstream utility assessment, and factuality checking in this study. 
As the results show, the synthetic clinical text generated by LLMs has both potential and weaknesses, and its usage should be assessed according to the application context. 

Two findings from our study warrant particular attention for practitioners using LLMs for clinical text generation. \textbf{The first is the long-context bottleneck.} The substantial ICD coding gap between note-level and chunk-level rephrasing suggests the use of chunking to improve generation quality. Meanwhile, the removal of surrounding context by chunking could lead to higher rates of hallucination.
This means \textit{a trade-off needs to be made on this decision based on the generative application and tolerance of each error type}, and future work needs to analyze in more depth the decisions related to chunking, such as boundary definition and chunk length. 

\textbf{Second, the error analysis shows that LLM failures are not random but exhibit recurring patterns.} This suggests the potential for \textit{future work to develop specialized techniques to address these errors}, such as better prompting for temporal anchoring and explicit handling of abbreviations. 

We used rephrasing as the generation method in this study, and we did not explicitly distinguish between ``rephrasing'' and ``synthesizing''. Nevertheless, we acknowledge the nuanced difference between them in healthcare, especially the connotation of the term ``synthesis'' for privacy preservation. Grounding on real notes to create alternative versions of the clinical text --- as done in ``rephrasing'' --- forfeits privacy protection. Meanwhile, since the goal of the study is to measure text quality and to shed light on LLMs' capabilities, we maintain this decision to allow evaluation in different setups, i.e., head-to-head comparison with human-written notes and on downstream tasks. 

Finally, our experiments show that the rephrased notes can be instrumental in understanding the quality of LLM generation and valuable for data augmentation despite being task-agnostic. To support future research, we compile and share the rephrased notes in this study as \dtitle, a synthetic corpus with 9B tokens based on 4.6M MIMIC notes.\footnote{The synthetic note corpus can be found at \url{https://github.com/JHLiu7/MIMIC-Ext-SynNotes}.}

\section*{Limitations}

\textbf{Choice of LLMs for generation.} We only focused on efficient open-source LLMs in our experiments.
While more recent models may handle long contexts better, recent work suggests the gap may be narrower than expected: \citet{Kang2025-vv} showed that using larger and more capable models for rephrasing pretraining data does not necessarily translate into better pretraining performance.
Our complementary results with Gemma and Mixtral (Appendix~\ref{sec:alter_model}) did not yield substantially different results, supporting this view.
Therefore, we consider the current model selection acceptable for fulfilling the study's objectives.
Nevertheless, extending the analysis to models at other scales --- both smaller and larger --- and considering frontier models in a privacy-preserved setting are important future work.

\textbf{Methods for text synthesis and evaluation.} As mentioned, given the scope of this study, we did not explore alternative approaches for generating synthetic clinical text, particularly reference-free methods. Since MIMIC contains abundant EHR data, future work could incorporate structured measurements in a cross-modal setting to enhance the quality and validity of synthetic text. For evaluation methodologies, we consider the use of models in the current study --- such as QuickUMLS for CUI extraction and BERT with sliding window for classification --- acceptable as they are consistent for synthetic and human-written notes, but future work needs to further conduct meta-evaluation of the validity and reliability of these methods. 

\textbf{Issues with fact-checking.} Although we conducted error analysis using LLM-as-judge for fact-checking, we were unable to directly compare synthetic notes with human-written notes. Such comparison requires extensive time and clinical expertise to review long notes and interpret patient care context, which was infeasible for this study. Instead, we opted to rely on downstream task performance and prior methods from fact-checking~\cite{Xie2024-uc} that demonstrate strong correlation with expert judgment. We also acknowledge the limitations of our current fact-checking setup, which is based on a single family of models and a small number of notes, as we presented it as an exploratory analysis. The choice of model for LLM-as-judge particularly needs further scrutiny, given that models can be sensitive in reported factuality scores~\cite{Munnangi2025-nc}. Future work needs to validate the robustness of alternative models --- probably the cheaper, open-source options --- for this critical task. Nevertheless, we kept the use of models and notes consistent throughout the experiment, so that the analysis of errors and the comparison between note-level and chunk-level rephrasing are still valid.

\section*{Ethical Considerations}

The generation of synthetic text using LLMs warrants careful ethical considerations. 
Although our synthesis was based on actual MIMIC notes and offered generally high precision of factual claims, we observed that synthetic notes still omit or introduce information, which can be further exacerbated by insufficient context. This highlights the need for cautious use and, in some cases, additional filtering to ensure reliability, depending on the use case. Furthermore, while we assessed potential bias in downstream models for risk prediction, other dimensions of bias might remain and need to be considered in future work.

We assessed the utility of synthetic notes in downstream modeling by comparing \textit{``train real, test real'' (TRTR)} and \textit{``train synthetic, test synthetic'' (TSTS)}. 
Real-world signals (e.g., diagnosis codes) were utilized in both settings, so the \textit{TSTS} models are applicable to \textit{``test real''}.
Meanwhile, careful consideration and transparency are essential if the synthetic notes were used to develop systems intended for implementation. 

We followed the recommended practices for handling MIMIC notes. For rephrasing, we used only local, open-source models, and conducted all experiments in a secure environment to ensure data protection. For fact-checking with Gemini, we adhered to PhysioNet's Responsible Use guidelines\footnote{https://physionet.org/news/post/gpt-responsible-use} by accessing Gemini models via Vertex AI, ensuring compliance with HIPAA regulations.

\section*{Acknowledgments}
We would like to thank our reviewers for their thoughtful, detailed, and constructive comments that helped to improve this paper. 
This research was supported in part by the Intramural Research Program of the National Institutes of Health (NIH). The contributions of the NIH author are considered Works of the United States Government. The findings and conclusions presented in this paper are those of the authors and do not necessarily reflect the views of the NIH or the U.S. Department of Health and Human Services.
This work was also supported by CSIRO (Australia’s National Science Agency) through the Australian e-Health Research Centre.

\bibliography{custom}

\appendix

\section{Details of Rephrasing MIMIC Notes}
\label{sec:generation}

For clinical text synthesis, each LLM was provided with a system prompt consistent with its chat template and was instructed to synthesize a new version using a paraphrasing prompt found effective in prior studies~\cite{Liu2024-iq,Maini2024-zd}, denoted as the ``high quality'' prompt:

\begin{promptbox}[Rephrasing prompt]
[System Prompt] You are a medical artificial intelligence assistant. The assistant gives truthful, detailed, and professional answers to the requests.

For the following paragraph give me a diverse paraphrase of the same in high-quality English language as in clinical notes written by medical professionals:
\end{promptbox}

\begin{table*}
\centering
\resizebox{.85\linewidth}{!}{
\begin{tabular}{l|c|c|c}
\hline
 & \# Total words (billion) & \# Avg words per note & \# Avg words per chunk \\ \hline
Llama (by notes) & 1.21 & 260 &  - \\
Mistral (by notes) & 1.34 & 289 &  - \\
Qwen (by notes) & 1.13 & 263 &  - \\
Llama (by chunks) & 2.24 & 482 & 206 \\
Mistral (by chunks) & 1.74 & 376 & 160 \\
Qwen (by chunks) & 1.41 & 315 & 146 \\
\hline
Total & 9.1 & 332 & - \\
\hline
\end{tabular}
}
\caption{Statistics of \dtitle generated by LLMs. When rephrasing based on chunks, each note from MIMIC is segmented into 2.3 chunks on average. The synthetic notes contain 9.1B words in total.}
\label{tab:synthetic}
\end{table*}

We consider two rephrasing setups, one \textit{by note} and the other \textit{by chunk}.
For rephrasing \textit{by note}, we stratify the clinical notes by lengths --- $(0, 500]$, $(500, 1000]$, $(1000, 2000]$, $(2000, 4000]$, and $>4000$ --- and set the corresponding maximum number of generated tokens to $1000$, $2000$, $4000$, $8000$, and $10{,}000$, respectively. For rephrasing \textit{by chunk}, we set the maximum number of generated tokens to $512$ given the short chunk length (\textasciitilde150 words).
Temperature was set to 0.75 and top-p 0.9 for all LLMs.

\section{Implementation Details on Extrinsic Evaluation}
\label{sec:eval-extrinsic}

We followed previous studies~\cite{Liu2022-eu} to construct the in-hospital prediction cohorts (mortality and DRG) from MIMIC-III. For ICD coding, we used the benchmark~\cite{Edin2023-zj} that comprehensively examined various types of ICD codes from MIMIC and focused on ICD-10 based on MIMIC-IV. Only ICD codes with a frequency higher than 10 were retained in the benchmark. This threshold was also applied to the DRG cohort with MS-DRG (Medicare Severity-DRG). We constructed a readmission cohort from MIMIC-IV by considering only the first admission of each patient and discarding immediate readmission cases that occurred within 24 hours after discharge. The statistics of the datasets are presented in Table \ref{tab:cohorts}.

We chose PLM-ICD~\cite{Huang2022-fn,Edin2023-zj} as the classification model given its strong performance on ICD coding and adaptability to handle long texts with sliding window. More specifically: 1) it leverages a pre-trained encoder (BERT), allowing for domain-specific initialization; 2) it processes text via a sliding-window mechanism, making it efficient for long clinical documents; 3) it has demonstrated strong performance, particularly for ICD coding, compared to other deep learning methods~\cite{Edin2023-zj}; and 4) it is adaptable across tasks by modifying only the classification layer.

We initialized the model with a biomedical RoBERTa checkpoint~\cite{Lewis2020-tw}. Training was conducted using a batch size of 16 for up to 10 epochs, with early stopping triggered after no improvement for 3 epochs. The learning rate was set to 5e-5, with 2,000 warmup steps followed by linear decay. This training recipe was applied to all versions of input notes, either human-written or LLM-rephrased, to ensure consistent comparisons. The trained model was then evaluated on the test set of the corresponding task. We averaged the results over five runs with different random seeds to report the main results.

\section{Fact-checking Implementation}
\label{sec:method-fact}

Following previous works~\cite{Munnangi2025-nc,Xie2024-uc}, we performed fact-checking of the synthetic notes by decomposing the synthetic notes into atomic facts and measuring entailment. 
We use Gemini-2.5-Pro, a thinking model, to decompose a note into atomic facts, and Gemini-2.5-Flash to assess entailment. We used the prompts from~\citet{Munnangi2025-nc} for fact decomposition and entailment judgment, which are presented in the following.

We sampled 25 notes of varied length (10 radiology reports and 15 discharge summaries) for fact-checking due to the cost related to querying Gemini. We also consider \textit{long notes} over 2K words, constituting five discharge summaries from the 25 notes. We show results (Table~\ref{tab:fact}) on both these 25 notes and the five long notes. For synthetic notes rephrased \textit{by chunk}, we also only present the results on the five long notes. Due to the prolonged content, these five notes take significantly longer to decompose and judge, sometimes requiring 500+ API calls for a single pair. Given the cost consideration, we relied on this small subset for this section.
We also only focused on Llama and Mistral for this analysis, which contributed 5.2K and 6.1K facts, respectively.

\begin{table*}[tbh]
\centering
\resizebox{0.85\linewidth}{!}
{%
\begin{tabular}{@{}lccccc@{}}
\toprule
  & \textbf{\# Patients} & \textbf{\% Train/Val/Test} & \textbf{\# Classes} & \textbf{Input notes} & \textbf{Metrics} \\
\midrule
In-hospital Mortality  & \makecell[t]{33,515\\(MIMIC-III)} & 80\%/10\%/10\%  & 2 (11.1\%) & \makecell[t]{Nursing notes,\\ physician notes, etc.}  & \makecell{AUROC\\ AUPRC} \\
30-Day Readmission    & \makecell[t]{33,518\\(MIMIC-IV)}  & 80\%/10\%/10\%  & 2 (13.0\%) & Discharge summary                                  & \makecell{AUROC\\ AUPRC} \\
DRG Prediction          & \makecell[t]{18,681\\(MIMIC-III)} & 70\%/20\%/10\%  & 369        & \makecell[t]{Nursing notes,\\ physician notes, etc.}  & \makecell{Accuracy\\ F1} \\
ICD Coding              & \makecell[t]{122,278\\(MIMIC-IV)} & 73\%/16\%/11\%  & 7,942      & Discharge summary                                  & \makecell{Micro F1\\ Macro F1} \\
\bottomrule
\end{tabular}%
}
\caption{Description of cohorts and prediction tasks. For diagnosis-related classifications, DRG prediction is based on MS-DRG and ICD coding on ICD-10. All tasks have a single label per patient except for ICD coding, which is multi-label multi-class classification.}
\label{tab:cohorts}
\end{table*}

\section{Omitted and Added Semantic Types}

Table~\ref{tab:semtype} presents the top-10 semantic types of the medical concepts that are either omitted or added by the LLMs after rephrasing. These are based on MedCON~\cite{Yim2023-jd} but mapped CUI to TUI. The added semantic types can be considered as false positives (FPs) and omitted ones as false negatives (FNs) when comparing against the human-written notes.

\section{Alternative Prompts}
\label{sec:alter_prompt}
We consider the following alternative prompts to explore their impact on generated text. 

\begin{description}
    \item[Rearranged Structure] \textit{Rewrite the following paragraph by rearranging sentences and phrases, while retaining the original clinical content and meaning:}
    \item[Minimal Changes] \textit{Rephrase the following paragraph with minimal changes, keeping the original structure and meaning intact. Make only slight adjustments for clarity and conciseness while preserving the clinical details:}
    \item[Formal Clinical Language] \textit{Paraphrase the following paragraph using strictly formal clinical documentation language, maintaining the same level of detail and medical accuracy:}
    \item[Substantial Revision] \textit{Produce a substantially revised version of the following paragraph in professional clinical English, changing sentence structure and vocabulary while preserving all critical medical details:}
    \item[Simpler English] \textit{Paraphrase the following paragraph into simpler clinical English, minimizing jargon and long sentences but keeping the key medical details:}
\end{description}

\begin{promptbox}[Entailment judgment prompt]
You are an expert on natural language entailment.
Your task is to deduce whether premise statements entail hypotheses.
Return only '1' if the hypothesis can be fully entailed by the premise.
Return only '0' if the hypothesis contains information that cannot be entailed by the premise.
Generate the answer in JSON format with the following keys:
'entailment_prediction': 1 or 0, whether the claim can be entailed.
Only return the JSON-formatted answer and nothing else.

Premise: {{premise}}
Hypothesis: {{hypothesis}}

Here is the JSON-formatted answer:

\end{promptbox}

\begin{promptbox}[Fact decomposition prompt]
Please breakdown the following text into independent facts as a string delimited by "//" to separate the facts

Example 1:
Note: "There is a dense consolidation in the left lower lobe."

Atomic facts:
There is a consolidation. // The consolidation is dense. // The consolidation is on the left. // The consolidation is in a lobe. // The consolidation is in the lower portion of the left lobe.

Example 2:

Note: "The patient has been having intermittent shortness of breath for the last two years."

Atomic facts:
The patient has been having shortness of breath. // The shortness of breath is intermittent. // The shortness of breath has been present for the last two years.

Do not include any other text, or say "Here is the list..."

Note: {{note}}
\end{promptbox}

\begin{table*}[h!]
    \centering
\resizebox{0.8\textwidth}{!}{
\begin{tabular}{ll}
\toprule
\multicolumn{2}{c}{\textbf{Synthetic text rephrased by Llama}} \\
\textit{Added Semantic Types (FP)} & \textit{Omitted Semantic Types (FN)} \\
\textit{Total counts: 9,321} & \textit{Total counts: 8,979} \\
\midrule
Natural Phenomenon or Process (19.1\%) & Medical Device (13.0\%) \\
Clinical Attribute (15.2\%) & Body Space or Junction (12.4\%) \\
Organ or Tissue Function (10.9\%) & Mental Process (11.7\%) \\
Pharmacologic Substance (9.9\%) & Biomedical or Dental Material (11.3\%) \\
Phenomenon or Process (8.1\%) & Gene or Genome (9.9\%) \\
Organism Attribute (7.9\%) & Body System (9.1\%) \\
Body Substance (7.6\%) & Indicator, Reagent, or Diagnostic Aid (8.8\%) \\
Organism Function (7.4\%) & Cell Component (8.5\%) \\
Laboratory or Test Result (7.3\%) & Organ or Tissue Function (7.9\%) \\
Body Part, Organ, or Organ Component (6.6\%) & Physiologic Function (7.3\%) \\
\end{tabular}
}

\resizebox{0.8\textwidth}{!}{
\begin{tabular}{ll}
\toprule
\multicolumn{2}{c}{\textbf{Synthetic text rephrased by Mistral}} \\
\textit{Added Semantic Types (FP)} & \textit{Omitted Semantic Types (FN)} \\
\textit{Total counts: 7,409} & \textit{Total counts: 4,238} \\
\midrule
Natural Phenomenon or Process (15.9\%) & Gene or Genome (14.6\%) \\
Clinical Attribute (11.9\%) & Body Space or Junction (13.0\%) \\
Phenomenon or Process (11.4\%) & Mental Process (11.3\%) \\
Organ or Tissue Function (11.2\%) & Medical Device (10.3\%) \\
Organism Attribute (9.9\%) & Cell Component (9.4\%) \\
Organism Function (8.5\%) & Biomedical or Dental Material (8.9\%) \\
Body Space or Junction (8.3\%) & Biologic Function (8.8\%) \\
Laboratory or Test Result (8.2\%) & Body System (8.0\%) \\
Pharmacologic Substance (7.8\%) & Indicator, Reagent, or Diagnostic Aid (7.9\%) \\
Body Substance (6.9\%) & Organ or Tissue Function (7.8\%) \\
\end{tabular}
}

\resizebox{0.8\textwidth}{!}{
\begin{tabular}{ll}
\toprule
\multicolumn{2}{c}{\textbf{Synthetic text rephrased by Qwen}} \\
\textit{Added Semantic Types (FP)} & \textit{Omitted Semantic Types (FN)} \\
\textit{Total counts: 5,641} & \textit{Total counts: 5,288} \\
\midrule
Natural Phenomenon or Process (13.1\%) & Mental Process (12.4\%) \\
Clinical Attribute (11.9\%) & Body Space or Junction (11.4\%) \\
Pharmacologic Substance (11.0\%) & Clinical Attribute (11.2\%) \\
Phenomenon or Process (10.4\%) & Gene or Genome (10.7\%) \\
Cell (9.8\%) & Medical Device (10.3\%) \\
Element, Ion, or Isotope (9.5\%) & Body Location or Region (9.3\%) \\
Laboratory or Test Result (8.8\%) & Nucleic Acid, Nucleoside, or Nucleotide (9.3\%) \\
Molecular Function (8.6\%) & Cell Component (8.6\%) \\
Organism Function (8.6\%) & Indicator, Reagent, or Diagnostic Aid (8.4\%) \\
Biomedical or Dental Material (8.3\%) & Organism Attribute (8.4\%) \\
\bottomrule
\end{tabular}
}
    \caption{Changes of semantic types in the rephrased notes by the three LLMs. Top 10 semantic types are listed for false positive (FP) and false negative (FN) compared to the original, human-written notes. The ratio is calculated based on total counts of the presented semantic types for each LLM for FP and FN, respectively.}
    \label{tab:semtype}
\end{table*}

\begin{table*}
    \centering
    \setlength{\tabcolsep}{6pt}
    \renewcommand{\arraystretch}{1.2}
    
\resizebox{0.93\textwidth}{!}{
    
    \begin{tabular}{p{0.40\textwidth} p{0.22\textwidth} >{\itshape\arraybackslash}p{0.30\textwidth}}
        \hline
        \textbf{Claim} & \textbf{Error Category} & \textbf{\upshape Comment} \\
        \hline

        ``The patient presented with fever.'' & No error & This is a correct claim. \\
        ``The patient continued to have significant oxygen requirements.'' & No error & Correct. \\

        ``An MRI later confirmed the focus was a subacute infarction.'' & Misinterpretation  & This is likely, but not confirmed. \\
        ``The anemia was treated with iron supplementation.'' & Fabricated claim & Patient had iron-binding study. No supplementation intake was documented. \\
        ``The surgery occurred on discharge day.'' & Misinterpretation  & Wrong interpretation of ``Surgery Discharge part''. \\
        ``The patient was cleared by occupational therapy (OT) for discharge to rehabilitation.'' & Misinterpretation  & Cleared by PT. \\
        ``The patient was scheduled for a follow-up appointment with her gastroenterologist.'' & Misinterpretation  & Appointment scheduled with primary care physician. \\

        ``The patient's name is Mr.\ John Doe.'' & Fabricated claim & Made-up name. \\
        ``The patient is a non-smoker.'' & Misinterpretation  & Incorrect; tobacco use is mentioned. \\
        ``The patient consumes approximately 1--2 drinks per day.'' & Measurement/Numeracy & EtOH: [**4--20**] drinks/day. \\

        ``The adjusted medications include simvastatin.'' & Fabricated claim & No adjustment. \\

        ``The patient is asymptomatic with the low blood sugar levels.'' & Misinterpretation  & ``He states that he is asymptomatic,'' but note shows symptomatic hypoglycemia. \\
        ``In the emergency department, the patient's vital signs were stable.'' & Misinterpretation  & Incorrect; labs showed elevated WBC. \\
        ``The patient's most recent pulse was 95\,bpm.'' & Temporal/Recency & Not most recent. \\
        ``The patient's most recent urine output was 2\,L.'' & Temporal/Recency & Not most recent. \\
        ``The patient is a former smoker.'' & Temporal/Recency & Current smoker. \\
        ``All pathology samples of bone and cartilage showed hyaline cartilage with focal acute inflammation.'' & Misinterpretation  & Incorrect summarization of pathology results: not all samples show hyaline cartilage with focal acute inflammation. \\

        ``In the ER, the patient's respiration rate was 142/min.'' & Measurement/Numeracy & 142 is blood pressure. \\
        ``In the ER, the patient's blood pressure was 18/97\,mmHg.'' & Measurement/Numeracy & 18 is respiratory rate; 96 (instead of 97) is oxygen saturation. \\

         ``The patient's discharge blood pressure was 168/45 mmHg.'' & Misinterpretation & BP in ED, not discharge. \\

        \hline
    \end{tabular}

    }
    \caption{Categorization and analysis of example false positive claims in synthetic texts.}
    \label{tab:errors}
    
\end{table*}

\section{Alternative Models}
\label{sec:alter_model}

Table~\ref{tab:performance-results} presents the numerical results presented in Figure~\ref{fig:extrinsic}. We provide additional results on rephrasing with Gemma-2 (9B) and Mixtral-v0.1 (8×7B) by chunks with one seed. 

\begin{table*}[tbh]
\centering
\resizebox{\textwidth}{!}{
\begin{tabular}{lcccccccc}
\toprule
 & \multicolumn{2}{c}{\textbf{In-hospital Mortality}} 
 & \multicolumn{2}{c}{\textbf{30-day Readmission}} 
 & \multicolumn{2}{c}{\textbf{ICD Coding}} 
 & \multicolumn{2}{c}{\textbf{DRG Classification}} \\
\cmidrule(lr){2-3}\cmidrule(lr){4-5}\cmidrule(lr){6-7}\cmidrule(lr){8-9}
\textbf{Input source} & \textbf{AUROC} & \textbf{AUPRC} & \textbf{AUROC} & \textbf{AUPRC} & \textbf{Micro F1} & \textbf{Macro F1} & \textbf{Accuracy} & \textbf{F1} \\
\midrule
Human-written
 & \makecell{0.873 \\ \small(0.858--0.888)}
 & \makecell{0.507 \\ \small(0.467--0.547)}
 & \makecell{0.627 \\ \small(0.566--0.688)}
 & \makecell{0.233 \\ \small(0.168--0.297)}
 & \makecell{0.564 \\ \small(0.562--0.566)}
 & \makecell{0.180 \\ \small(0.172--0.189)}
 & \makecell{0.314 \\ \small(0.308--0.321)}
 & \makecell{0.132 \\ \small(0.115--0.149)} \\
 
\midrule

\multicolumn{9}{l}{Synthetic text \textit{by notes}} \\[2pt]

Llama
 & \makecell{0.874 \\ \small(0.853--0.895)}
 & \makecell{0.521 \\ \small(0.480--0.561)}
 & \makecell{0.602 \\ \small(0.514--0.691)}
 & \makecell{0.219 \\ \small(0.149--0.289)}
 & \makecell{0.458 \\ \small(0.456--0.460)}
 & \makecell{0.138 \\ \small(0.131--0.144)}
 & \makecell{0.306 \\ \small(0.301--0.311)}
 & \makecell{0.125 \\ \small(0.106--0.144)} \\[6pt]

Mistral
 & \makecell{0.879 \\ \small(0.871--0.888)}
 & \makecell{0.528 \\ \small(0.502--0.555)}
 & \makecell{0.637 \\ \small(0.609--0.664)}
 & \makecell{0.241 \\ \small(0.212--0.269)}
 & \makecell{0.522 \\ \small(0.521--0.524)}
 & \makecell{0.158 \\ \small(0.154--0.161)}
 & \makecell{0.315 \\ \small(0.311--0.318)}
 & \makecell{0.138 \\ \small(0.125--0.150)} \\[6pt]

Qwen
 & \makecell{0.861 \\ \small(0.849--0.872)}
 & \makecell{0.477 \\ \small(0.447--0.508)}
 & \makecell{0.564 \\ \small(0.500--0.628)}
 & \makecell{0.182 \\ \small(0.119--0.244)}
 & \makecell{0.452 \\ \small(0.447--0.458)}
 & \makecell{0.122 \\ \small(0.110--0.135)}
 & \makecell{0.299 \\ \small(0.295--0.303)}
 & \makecell{0.124 \\ \small(0.108--0.139)} \\[6pt]

\midrule

\multicolumn{9}{l}{Synthetic text \textit{by chunks}} \\[2pt]
 
Llama
 & \makecell{0.874 \\ \small(0.857--0.892)}
 & \makecell{0.521 \\ \small(0.464--0.579)}
 & \makecell{0.654 \\ \small(0.626--0.683)}
 & \makecell{0.265 \\ \small(0.237--0.293)}
 & \makecell{0.539 \\ \small(0.538--0.541)}
 & \makecell{0.172 \\ \small(0.161--0.183)}
 & \makecell{0.305 \\ \small(0.298--0.312)}
 & \makecell{0.127 \\ \small(0.119--0.136)} \\[6pt]
 
Mistral
 & \makecell{0.881 \\ \small(0.872--0.890)}
 & \makecell{0.530 \\ \small(0.504--0.555)}
 & \makecell{0.643 \\ \small(0.596--0.690)}
 & \makecell{0.244 \\ \small(0.187--0.301)}
 & \makecell{0.549 \\ \small(0.547--0.551)}
 & \makecell{0.177 \\ \small(0.167--0.187)}
 & \makecell{0.309 \\ \small(0.305--0.314)}
 & \makecell{0.133 \\ \small(0.118--0.149)} \\[6pt]

Qwen
 & \makecell{0.868 \\ \small(0.862--0.874)}
 & \makecell{0.492 \\ \small(0.458--0.526)}
 & \makecell{0.618 \\ \small(0.552--0.683)}
 & \makecell{0.229 \\ \small(0.161--0.296)}
 & \makecell{0.519 \\ \small(0.517--0.521)}
 & \makecell{0.162 \\ \small(0.153--0.170)}
 & \makecell{0.293 \\ \small(0.290--0.296)}
 & \makecell{0.121 \\ \small(0.101--0.142)} \\
 
\midrule

\multicolumn{9}{l}{Additional synthesis \textit{by chunks}} \\[2pt]
Gemma & 0.870 & 0.512 & 0.654 & 0.285 & 0.537 & 0.171 & 0.293 & 0.109 \\
Mixtral & 0.882 & 0.548 & 0.662 & 0.275 & 0.548 & 0.169 & 0.314 & 0.123 \\
\bottomrule
\end{tabular}}
\caption{Downstream modeling results for the four tasks based on different sources of clinical notes. Values are shown as mean with 95\% confidence intervals in parentheses. The results for the three main LLMs are also presented in Figure~\ref{fig:extrinsic}.}
\label{tab:performance-results}
\end{table*}

\end{document}